# German Dialect Identification Using Classifier Ensembles

**Alina Maria Ciobanu[1], Shervin Malmasi[2], Liviu P. Dinu[1]**
[1]University of Bucharest, Romania
[2]Harvard Medical School, United States
alina.ciobanu@my.fmi.unibuc.ro

## Abstract

In this paper we present the GDI_classification entry to the second German Dialect Identification (GDI) shared task organized within the scope of the VarDial Evaluation Campaign 2018. We present a system based on SVM classifier ensembles trained on characters and words. The system was trained on a collection of speech transcripts of five Swiss-German dialects provided by the organizers. The transcripts included in the dataset contained speakers from Basel, Bern, Lucerne, and Zurich. Our entry in the challenge reached 62.03% F1 score and was ranked third out of eight teams.

## 1 Introduction

Discriminating between dialects and language varieties is a challenging aspect of language identification that sparked interest in the NLP community in the past few years. As evidenced in a recent survey (Jauhiainen et al., 2018), a number of papers have been published on this topic on dialects of Arabic (Tillmann et al., 2014) and Romanian (Ciobanu and Dinu, 2016), and language varieties of English (Lui and Cook, 2013) and Portuguese (Zampieri et al., 2016).

This challenge motivated the organization of a number of competitions such as the Discriminating between Similar Languages (DSL) shared tasks which included language varieties and similar languages (Zampieri et al., 2014; Zampieri et al., 2015; Malmasi et al., 2016b), the MGB challenge 2017 (Ali et al., 2017) on Arabic, the PAN lab on author profiling, which in 2017 included varieties and dialects (Rangel et al., 2017), and finally the first German Dialect Identification (GDI) shard task in 2017 (Zampieri et al., 2017). The GDI shared task 2017 preceded the second GDI shared task (Zampieri et al., 2018) in which our team, GDI_classification, participated.

In this paper we describe the GDI_classification system trained to identify four dialects of (Swiss) German. The GDI dataset included speech transcripts from speakers from Basel, Bern, Lucerne, and Zurich. The system is based on an ensemble of multiple SVM classifiers trained on words and characters as features. Our approach is inspired by the approach of Malmasi and Zampieri (2017b), which was ranked first in the first edition of the GDI task and also performed well on identifying dialects of Arabic (Malmasi and Zampieri, 2017a). We build on the experience of previous work of members of the GDI_classification team improving a system that we have previously applied to a similar classification task, namely author profiling (Ciobanu et al., 2017).

## 2 Related Work: The First GDI Shared Task

There have been a few studies on German dialect identification published before the first GDI shared task, using different corpora and evaluation methods (Scherrer and Rambow, 2010; Hollenstein and Aepli, 2015). To the best of our knowledge, the first GDI shared task organized in 2017 was the first attempt to provide a benchmark for this task.

The GDI shared task 2017 setup and dataset were similar to those of the 2018 edition presented in more detail in Section 3. The results of the 2017 edition along with a reference to each system description paper are presented in Table 1.

| Rank | Team | F1 (weighted) | Reference |
|---|---|---|---|
| 1 | MAZA | 0.662 | (Malmasi and Zampieri, 2017b) |
| 2 | CECL | 0.661 | (Bestgen, 2017) |
| 3 | CLUZH | 0.653 | (Clematide and Makarov, 2017) |
| 4 | qcri_mit | 0.639 | - |
| 5 | unibuckernel | 0.637 | (Ionescu and Butnaru, 2017) |
| 6 | tubasfs | 0.626 | (Çöltekin and Rama, 2017) |
| 7 | ahaqst | 0.614 | (Hanani et al., 2017) |
| 8 | Citius_Ixa_Imaxin | 0.612 | (Gamallo et al., 2017) |
| 9 | XAC_Bayesline | 0.605 | (Barbaresi, 2017) |
| 10 | deepCybErNet | 0.263 | - |

Table 1: GDI shared task 2017: Closed submission results.

The ten teams who competed in the first GDI challenge applied different computational methods to approach the task. These include linear SVM classifiers (Çöltekin and Rama, 2017; Bestgen, 2017), string kernels (Ionescu and Butnaru, 2017), Naive Bayes classifiers (Barbaresi, 2017), and SVM ensembles (Malmasi and Zampieri, 2017b), which achieved the first place in 2017. For this reason, this is the approach we apply in our GDI_identification system.

## 3 Data

In this paper we used only the dataset provided by the GDI organizers. The dataset is part of the ArchiMob corpus (Samardžić et al., 2016).[1] It contains transcripts of interviews with speakers from Basel (BS), Bern (BE), Lucerne (LU), and Zurich (ZH). The interviews have been transcribed using the 'Schwyzertütschi Dialäktschrift' system (Dieth, 1986).

The evaluation was divided into two tracks. In the first of them organizers provided participants with a test set containing the four aforementioned dialects included in the training set. In the second track they provided a test set containing the four dialects plus a 'surprise' dialect not included in the training set. We opted to participate only in the first track which contained only previously 'seen' dialects.

The dataset comprise nearly 25,000 instances divided in training, development, and test partitions as presented in Table 2.

| Partition | Instances |
|---|---|
| Training | 14,647 |
| Development | 4,659 |
| Test | 5,543 |
| **Total** | **24,849** |

Table 2: Instances in the GDI dataset 2018.

## 4 Methodology

The system that we propose for the GDI shared task consists of an ensemble of classifiers, namely SVMs. In this approach, we employ the methodology proposed by Malmasi and Dras (2015).

Ensembles of classifiers are deemed useful when there are disagreements between the comprising classifiers, which can use different features, training data, algorithms or parameters. The scope of the ensemble is to combine the results of the classifiers in such a way that the overall performance is improved over the individual performances of the classifiers. Ensembles have proven useful in various tasks, such as complex word identification (Malmasi et al., 2016a) and grammatical error diagnosis (Xiang et al., 2015).

---
[1] http://www.spur.uzh.ch/en/departments/research/textgroup/ArchiMob.html

To distinguish the classifiers, we employ a different type of features for each of them. After obtaining predictions from each classifier, they need to be combined, to obtain the final predictions of the ensemble. To implement this system we used the Scikit-learn (Pedregosa et al., 2011) library. For the individual classifiers, we used LinearSVC,[2] an SVM implementation based on the Liblinear library (Fan et al., 2008), with a linear kernel. For the ensemble, we used the VotingClassifier,[3] with a majority rule fusion method: for each instance, the class that has been predicted by the majority of the classifiers is considered the final prediction of the ensemble. When there are ties, this implementation chooses the final prediction based on the ascending sort order of all labels.

### 4.1 Features

Each classifier from the ensemble uses one of the following features with TF-IDF weighting:

- Character $n$-grams, with $n$ in $\{1, ..., 8\}$;

- Word $n$-grams, with $n$ in $\{1, 2, 3\}$;

- Word $k$-skip bigrams, with $k$ in $\{1, 2, 3\}$.

We obtain, thus, 14 classifiers. Training the individual classifiers, we achieve the results reported in Table 3. The features that lead to the best results are character 4-grams. The SVM using these features obtains 0.621 F1 score during our evaluation on the development dataset. In Table 4 we report the most informative character 4-grams for each class.

| Feature | F1 (macro) |
|---|---|
| Character 1-grams | 0.349 |
| Character 2-grams | 0.569 |
| Character 3-grams | 0.618 |
| **Character 4-grams** | **0.621** |
| Character 5-grams | 0.604 |
| Character 6-grams | 0.583 |
| Character 7-grams | 0.555 |
| Character 8-grams | 0.509 |
| Word 1-grams | 0.617 |
| Word 2-grams | 0.529 |
| Word 3-grams | 0.381 |
| Word 1-skip bigrams | 0.532 |
| Word 2-skip bigrams | 0.541 |
| Word 3-skip bigrams | 0.544 |

Table 3: Classification F1 score for individual classifiers on the development dataset.

To improve the performance of the SVM classifier using character 4-grams as features, we built various ensembles and performed a grid search with the purpose of determining the optimal value for the SVM regularization parameter $C$. We searched in $\{10^{-3}, ..., 10^3\}$ and obtained the optimal value 1. Experimenting with different classifiers as part of the ensemble, we determined the optimal feature combination to be character $n$-grams with $n$ in $\{2, 3, 4, 5\}$. This ensemble obtained 0.638 F1 score on the development dataset.

---

[2] http://scikit-learn.org/stable/modules/generated/sklearn.svm.LinearSVC.html
[3] http://scikit-learn.org/stable/modules/generated/sklearn.ensemble.VotingClassifier.html

| Class | Most Informative Features |
|-------|---------------------------|
| BE    | \| näc\| aus \|seit\|nei \|gsee\| ein\| ds \| aus\|hei \| hei\| |
| BS    | \| uns\|aabe\| kha\|rlig\| kho\| äu \| häi\| dr \|häi \| net\| |
| LU    | \| sen\| hem\| gch\|ech \| oü \| hai\|gcha\|hend\|ond \| hen\| |
| ZH    | \|maal\|zwäi\|deet\|änn \|dän \| dän\| hät\|dänn\|hät \| näd\| |

Table 4: Top 10 most informative character 4-grams for each class.

## 5 Results

We submitted a single run for the GDI task for the official evaluation. The results of our system and of a random baseline (provided by the organizers) on the test dataset are reported in Table 5. The run that we submitted corresponds to our best performing SVM ensemble, comprising 4 classifiers, each using one of the following features: character $n$-grams with $n$ in $\{2, 3, 4, 5\}$. Our system was ranked third, obtaining 0.6203 F1 score on the test set, significantly outperforming the random baseline, which obtained 0.2521 F1 score on the test set.

In the official evaluation, the ranking was made taking statistical significance into account, and thus our team ranked third. The best performing team obtained 0.685 F1 score on the test dataset.

| System          | F1 (macro) |
|-----------------|------------|
| Random Baseline | 0.2521     |
| **SVM Ensemble**| **0.6203** |

Table 5: Results for the GDI shared task on the test dataset.

Looking at the confusion matrix of our system (see Figure 1), we notice that BS is identified correctly most often, while LU is at the opposite end, with the lowest number of correctly classified instances. Out of the misclassified instances, we noticed that LU sentences are very often identified as BE.

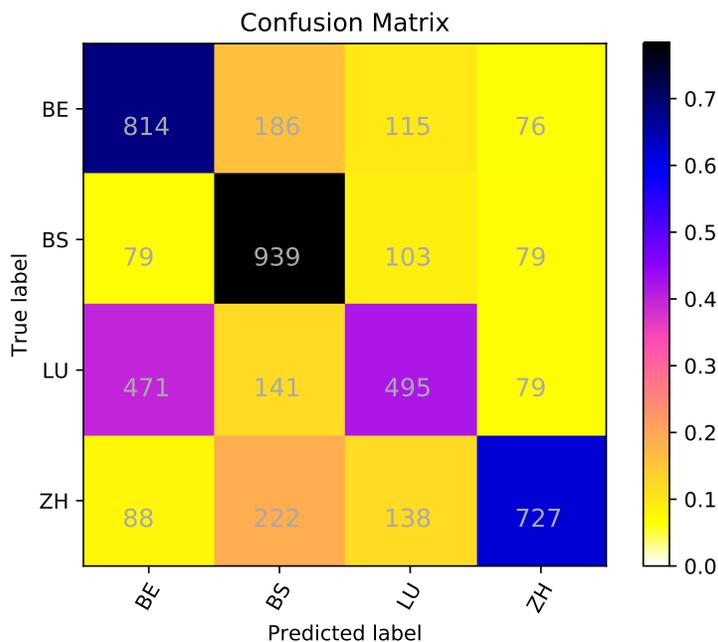

Figure 1: Confusion matrix for the SVM ensemble on the GDI shared task. The four Swiss German dialects are abbreviated as follows: Bern (BE), Basel (BS), Lucerne (LU) and Zurich (ZH).

## 6 Conclusion

In this paper we presented the GDI_identification entry for the GDI shared task at VarDial 2018. We employed an ensemble of SVM classifiers, continuing our previous work in this direction (Ciobanu et al., 2017). We obtained 0.6203 F1 score on the test dataset, ranking third in the competition. We experimented with various character and word n-gram features, and obtained our best performance with an ensemble of four classifiers, each using a different group of features. A variation of this system has been submitted for the Indo-Aryan language identification (ILI) shared task at VarDial 2018 (Ciobanu et al., 2018), achieving good performance, ranking third out of eight teams.

As future work, we intend to improve our ensemble implementation and to experiment with other features as well as in Bestgen (2017), in order to improve the performance of our system.

## Acknowledgements

We would like to thank the GDI organizers, Yves Scherrer and Tanja Samardžić, for organizing the shared task and for replying promptly to our questions.

We further thank the anonymous reviewers and Marcos Zampieri for the feedback and suggestions provided.